\documentclass[conference, compsoc]{IEEEtran}
\IEEEoverridecommandlockouts

\usepackage[utf8]{inputenc} 
\usepackage[T1]{fontenc}    
\usepackage{url}            
\usepackage{booktabs}       
\usepackage{amsfonts}       
\usepackage{nicefrac}       
\usepackage{microtype}      
\usepackage[numbers]{natbib}

\usepackage{amsmath}
\usepackage{amssymb}
\usepackage{caption}
\usepackage{subcaption}
\usepackage{wrapfig}
\usepackage{algorithmic}
\usepackage{algorithm,comment}

\usepackage{graphicx}
\usepackage{bmpsize}

\usepackage{amsmath,mathtools}
\usepackage{lastpage,fancyhdr}

\title{EEG to fMRI Synthesis:\break Is Deep Learning a candidate?}

\author{%
  David Calhas\\
  Instituto Superior Técnico\\
  Lisbon, Portugal\\
  \texttt{david.calhas@tecnico.ulisboa.pt}\\
  \and
  Rui Henriques \\
  Instituto Superior Técnico \\
  Lisbon, Portugal \\
  \texttt{rmch@tecnico.ulisboa.pt}\\
}

\begin{document}

\maketitle

\begin{abstract}
Advances on signal, image and video generation underly major breakthroughs on generative medical imaging tasks, including Brain Image Synthesis. Still, the extent to which functional Magnetic Ressonance Imaging (fMRI) can be mapped from the brain electrophysiology remains largely unexplored. This work provides the first comprehensive view on how to use state-of-the-art principles from Neural Processing to synthesize fMRI data from electroencephalographic (EEG) data. Given the distinct spatiotemporal nature of haemodynamic and electrophysiological signals, this problem is formulated as the task of learning a mapping function between multivariate time series with highly dissimilar structures. A comparison of state-of-the-art synthesis approaches, including Autoencoders, Generative Adversarial Networks and Pairwise Learning, is undertaken. Results highlight the feasibility of EEG to fMRI brain image mappings, pinpointing the role of current advances in Machine Learning and showing the relevance of upcoming contributions to further improve performance. EEG to fMRI synthesis offers a way to enhance and augment brain image data, and guarantee access to more affordable, portable and long-lasting protocols of brain activity monitoring. The code used in this manuscript is available in Github and the datasets are open source.
\end{abstract}

\section{Introduction}\label{section:introduction}


\begin{figure}[ht]
    \centering
    \begin{subfigure}[b]{0.23\textwidth}
        \centering
        \includegraphics[width=\textwidth]{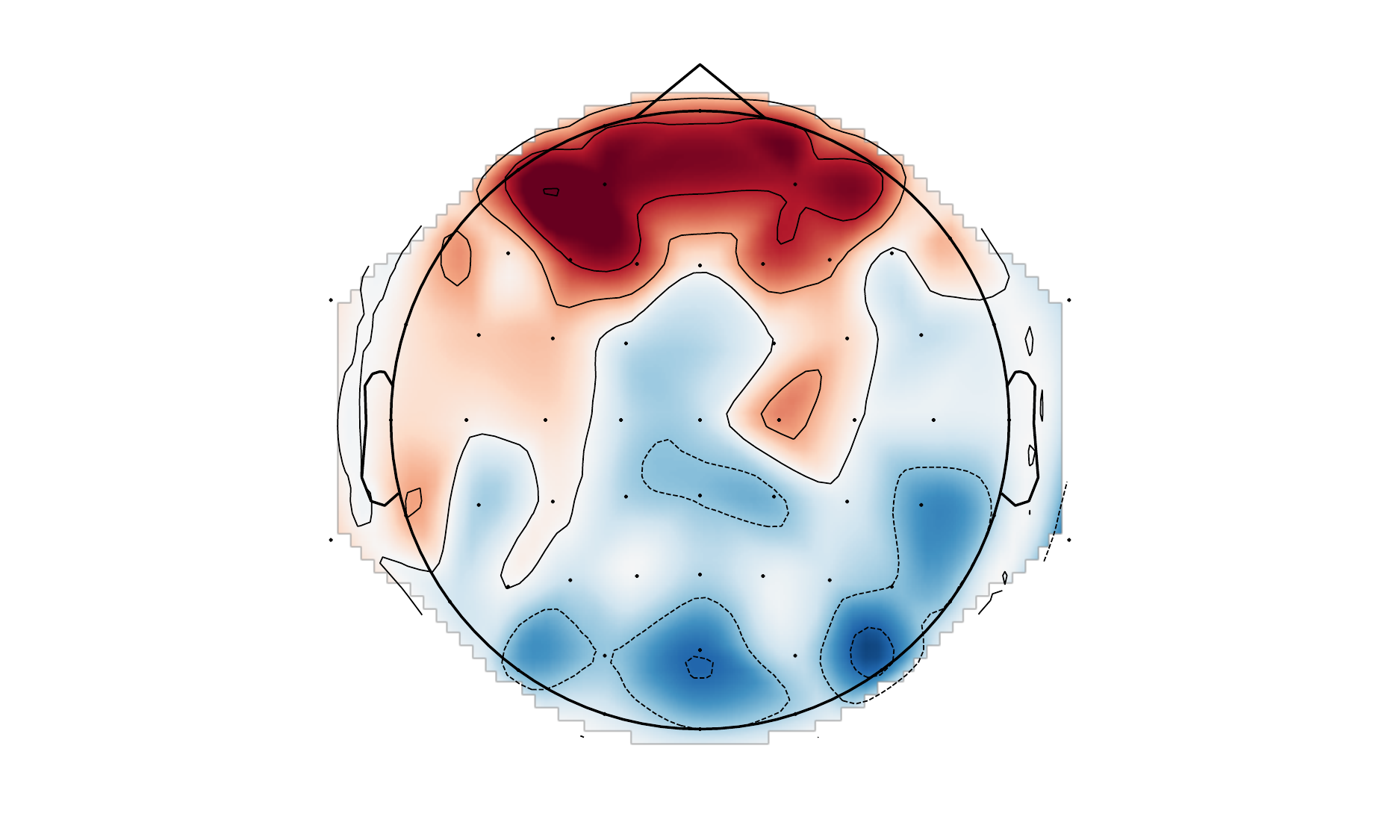}
        \caption{EEG source localization topography map}
        \label{fig:eeg_source_localization}
    \end{subfigure}
    \hfill
    \begin{subfigure}[b]{0.23\textwidth}
        \centering
        \includegraphics[width=0.7\textwidth, height=3cm]{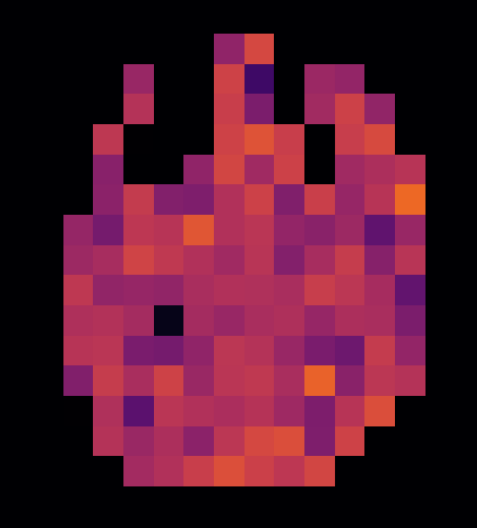}
        \caption{fMRI sliced at $z$ axis}
        \label{fig:fmri}
    \end{subfigure}
    \caption{\small Simplified spatial representation of electrophysiological and haemodynamic signals simultaneously collected from an individual of the NODDI dataset (Section \ref{subsection:dataset_description}), showing that EEG and fMRI data yield distinct properties. EEG signals from $64$ channels are taken to a spatial representation using source localization techniques \citep{gramfort2013mne}.}
    \label{fig:comparison_eeg_source_fmria}
\end{figure}

Signal Generation approaches explore how structural transformations can be learned from a given signal collection to encode and/or produce new signals. 
When the goal is to learn a mapping function between signals 
from heterogeneous sources, the focus is placed on translations between those sources, transferring the task from Generation to Synthesis. 
Recent breakthroughs on brain image generation \citep{abramian2019generating, chao2018generating}, reconstruction \citep{aggarwal2016accelerated}, enhancement \citep{He_2019_CVPR} and synthesis \citep{ben2019cross} 
are driven by the simultaneous analysis of multiple imaging modalities, mostly Computed Tomography (CT), functional Magnetic Resonance Imaging (fMRI) and Positron Emission Tomography (PET) \citep{YI2019101552}. Yi et al. \cite{YI2019101552} survey recent works that establish mappings between CT, fMRI and PET modalities. In spite of the increasing number of contributions, a noticeable lack in the existing research is the absence of mappings between electroencephalography (EEG) and fMRI data. A contributor factor is the inherent difficulty of mapping electrophysiological and haemodynamic signals, given their contrasting spatial and temporal resolution (Figure \ref{fig:comparison_eeg_source_fmria}). 
Nevertheless, the importance of this task has been largely evidenced:
\begin{enumerate}
\item  MRI units are still largely scarce in countries worldwide \citep{mikulic_2019}. \cite{ogbole2018survey} estimate the presence of $0.24$ units per million people in West African countries; 
\item  in contrast with other brain imaging modalities, electroencephalography is non-invasive, safe, inexpensive, and yields almost no 
restriction on the extent of recordings \citep{fowle2000uses}. Research and medical-wise, EEG to fMRI synthesis opens up the possibility to perform more affordable, portable and long-lasting protocols of brain activity monitoring;
\item simultaneous EEG and fMRI monitoring provides a way of complementing strengths and addressing the limitations of both signals: EEG offers a fine temporal and spectral resolution of the brain electrophysiology, while fMRI offers a precise spatial resolution of blood flow changes (associated with brain activity) \citep{murta2015electrophysiologicalco, zhu2019100855, labounek2019eegsp};
\item EEG-based generation of new fMRI images can be further used as a way of augmenting data or guaranteeing its proper privacy \citep{zhuang2019fmri};
\end{enumerate}

Understanding the extent to which these modalities can be mapped is critical to answer key research problems:

\begin{itemize}
\item unravel the complex neurophysiological relationships between the brain's cortical electrophysiology and its haemodynamic;
\item access the components of EEG signals that are decodable and non-decodable into fMRI signals, unraveling their role;
\item identify the brain regions from each modality that support the synthesis process. This knowledge can be, for instance, used to reveal the semantics of brain activity and its underlying connectivity.
\end{itemize}

This manuscript proposes an approach to synthesize fMRI from EEG signals based on the composition of convolution (encoding) and transposed convolution (decoding) layers. Given the rich spatiotemporal nature of both modalities, this problem is formulated as learning a mapping function between multivariate time series with highly dissimilar structures (regarding both spatial and temporal resolution). To this end, we provide a comprehensive comparison of state-of-the-art principles from Neural Processing towards multivariate time series data analysis, together with well-established principles for the integrative analysis of EEG and fMRI data:
\begin{enumerate}
\item Autoencoders (AE) \cite{rumelharthinton1986ae} as a baseline;
\item Generative Adversarial Networks (GAN) \cite{goodfellow2014generativean} with the entropy and Wasserstein (WGAN) losses;
\item linear combination of top-$k$ most correlated fMRI-EEG training data instances; 
\item novel class of neural networks combining contras- tive loss \cite{contrastive2005chopra} at the encoder level with reconstruction loss at the decoder level (details in Section \ref{section:approach}).
\end{enumerate}

\noindent Results show, on one hand, the feasibility of EEG-based synthesis of fMRI signals, pinpointing the role of 
current advances on the field to tackle this challenging task. On the other hand, they highlight a significant space to improve performance, stressing the relevance of upcoming contributions to the targeted task. 

This article is organized as follows. Section \ref{section:related_work} provides essential background on deep generative models and relevant work on brain imaging synthesis. Section \ref{section:dataset} describes the datasets and preprocessing protocols considered in the context of our study. Section \ref{section:approach} introduces the proposed approaches, covering principles on how to synthesize BOLD signals 
from EEG signals. Section \ref{section:results} discusses the gathered empirical results. Finally, concluding remarks and future directions are presented. 

\section{Related Work}\label{section:related_work}

\vskip -0.1cm
\noindent Image Reconstruction and Image Synthesis and their \textit{unconditional}, \textit{cross-modal} and \textit{constrained} variants are important tasks towards our end. On Image Reconstruction, learning methods typically establish transformations to 
reduce noise, remove artifacts and produce descriptors \citep{zhuang2019fmri, turek2017semi}. These approaches can be extended for medical Image Synthesis, where the goal is to learn transformations for auto-encoding signals of medical nature (MRI, fMRI, EEG, CT, PET). Cross-Modality Image Synthesis maps raw signals from one modality, such as magnetic resonance images, onto signals from another modality, such as CT-like images. This cross-modal translation can be achieved in the presence of both \textit{paired} images (multimodal recordings) and \textit{unpaired} images (non-simultaneous  recordings). Understandably, unpaired instances do not respect the data fidelity loss term, being unable to preserve small abnormality regions during the translation process. Unconditional Image Synthesis, also known as Image Generation, learns transformations on samples taken from a certain distribution to generate images resembling the ones considered in the learning phase \citep{chen2016variational}. Constrained Image Synthesis has its base on applying transformations respecting constraints on the modality being synthesized, such as segmentation maps (e.g. suppressing bones, learning mappings between regions of interest). 

The interest and necessity to perform synthesis between brain imaging modalities has been largely motivated \citep{YI2019101552}. \citet{dong2017} inferred CT images from their corresponding MR images using adversarial training from a fully convolutional network. Similar networks are used by \citet{wolterink2017} to map 2D brain MR image slices into a 2D brain CT image. \citet{YI2019101552} provide an extensive survey on multi-modal brain image synthesis from a stance of adversarial training.

Despite the existing advances on this field, to our knowledge there are not comprehensive attempts to answer 
EEG to fMRI synthesis. Two observations may explain this observation. First, the still scarce access to simultaneous EEG and fMRI scans. Second, the difficulty of establishing mappings between these two modalities given their highly distinct spatiotemporal dynamics. In particular, unlike the promising role that adversarial training has in the context of aforementioned studies on multi-modality image synthesis, the principles brought forth by adversarial training are indeed insufficient to deal with the complexity of associations between EEG and fMRI signals (results in Section \ref{section:results}).

In this work, we use contributions from Cross-Modality Image Synthesis tackle the task of synthesizing fMRI data from EEG data using paired recordings (simultaneous EEG and fMRI signals). Section \ref{relnp} introduces essential neural processing principles for Image Synthesis, while section \ref{rel:eegmri} surveys state-of-the-ark work on simultaneous EEG and fMRI studies. 

\subsection{Neural Processing for Image Synthesis}
\label{relnp}
\vskip -0.15cm
\noindent Our work builds upon recent deep learning techniques to synthesize multivariate time series, being inspired on: AE \cite{hinton2006reducing}, Variational AEs (VAE) \cite{kingma2013autoencodingvb}, $\beta$-VAE \cite{higgins2017betavae}, GAN \cite{goodfellow2014generativean}, WGAN \cite{arjovsky2017wassersteing} and Conditional GANs (CGAN) \cite{mirza2014conditionalga}; with some tweaks to each version in order to adapt them to the task at hand (EEG to fMRI synthesis). Since the goal is to synthesize and not to perform signal generation, samples from distributions are not taken (as it happens with VAE, $\beta$-VAE, GAN and WGAN). Instead, similar to CGANs, a decoder synthesizes fMRI based on a hidden representation of the EEG signal (with no concatenation of a random sample). This technique is also known as style transfer in GANs \citep{karrasla19, reed2016generativeat}. 
In fact, cross-modality image synthesis is one of the most important application of GANs \citep{YI2019101552}. Magnetic reasonance (MR) is ranked as the top medical imaging modality explored in GAN-related literature \citep{YI2019101552}, given the current costs and constraints on MR acquisition. GANs hold the potential to reduce MR acquisition time by faithfully generating sequences from already acquired ones. However, the bounds on the available data and convergence difficulties limit their success.

In cross-modality image synthesis, there are not yet reference loss functions and final metrics for assessing the generative accuracy of the models \citep{YI2019101552}. Most works opt to use traditional distance metrics such as Mean Absolute Error (MAE), Peak Signal-to-Noise Ratio (PSNR), or Structural Similarity (SSIM) for quantitative evaluation \citep{hore2010image}. These measures, however, do not always correspond to the visual quality of the image and disregard time dependencies along image frames. Therefore, additional metrics are proposed in Section \ref{approach:evaluation} and, in addition to quantitative results, qualitative results are complementarily presented in Section \ref{section:results}.


\subsection{Simultaneous EEG and fMRI studies}
\label{rel:eegmri}

\noindent\citet{he2018spatialtemporaldo} performed a simultaneous EEG and fMRI study to take advantage of both temporal and spatial precision of the EEG and fMRI, respectively. Their work explores the integration between gesture and speech under a thorough analysis between alpha and beta power and BOLD. Results suggest positive correlation between BOLD and alpha power and show that the temporal resolution for spectral content affect the strength of associations. This work leaves open questions, as it reduces electrophysiology to alpha and beta bands.

\citet{chang2013eeg} collected simultaneous EEG-fMRI data under a 
resting state condition from 10 healthy adults, and examine whether temporal variations in pairwise coupling of functional connectivity networks (based on fMRI) are associated with temporal variations in the amplitude of EEG power, specifically on alpha and theta frequency bands.
Functional connectivity networks were defined using an atlas of functional regions of interest that had been defined from a group-level independent component analysis of resting state fMRI. 
Decreases in alpha and increases in theta over time were associated with relative increases in functional connectivity. Positive correlations with alpha power were also observed in the thalamus and dorsal anterior cingulate cortex. Although these results motivate the possibility to establish EEG to fMRI mappings, they are constrained to specific spectral bands and connectivity maps, neglecting the rich nature of the EEG and haemodynamic signals. 

\citet{leite2013transfer} explore different EEG-fMRI transfer functions. For this purpose, metrics extracted from the EEG spectrum (under Morlet wavelet spectral analysis) were associated with haemodynamics for a single epileptic subject.
Significant correlations were reported. 
Yet, 
the lack of observations and the peculiar electrophysiology of epileptic subjects hamper the target learning of EEG-fMRI transfer functions. 
Similarly, \citet{rosa2010estimating} estimated EEG-fMRI transfer functions finding changes in BOLD associated with changes in the EEG spectrum. According to them, these changes do not arise from one specific band, but from the relative power of high and low frequencies. This shows how previous studies \citep{chang2013eeg,he2018spatialtemporaldo} would possibly improve results by exploring more frequency bands.

\citet{cury2019sparse} 
predict combined EEG and fMRI neurofeedback (NF) scores from EEG NF scores. 
The main goal was to perform a real time NF session using only EEG recording, instead of the costly and non-portable fMRI sessions. 
The dataset used consisted on a group of 17 subjects. The EEG recording was performed with 64 channels and sampled at 5kHz, while fMRI recordings were produced from a 3 Tesla scanner. 
The best approach on the training (testing) set claims a Pearson Correlation with mean $0.82$ ($0.74$), an improvement of 10pp against the baseline EEG NF scores. 
The model is elegant, and more transformations could be added (go from perceptrons to multi-perceptrons) as chains, which is the same as saying deep learning could improve results. 
In contrast, our work aims at synthesizing BOLD from EEG signals, instead of predicting extracted features (NF scores) from both modalities.

\citet{wei2020bayesian} is another study that complements fMRI signal with EEG information using Bayesian fusion. They compare the single  use of fMRI signal against complementing fMRI with EEG using a Bayesian belief updating, measuring the added value of EEG. 
\citet{mosayebi2020correlated} also perform EEG and fMRI fusion by means of a matrix factorization algorithm called Correlated Coupled Matrix Tensor Factorization 
forcing EEG and fMRI to share the same feature space. The results reported show that there is not a consistent correlation among the extracted features (please check the original work for more details).
\citet{jiang2020targeting} synthesize a functional transcranial brain atlas. Although, they do not perform an actual modality synthesis, this is another example of the upwards trend of the functional neuroimaging modalities synthesis research.

\section{EEG-fMRI data}\label{section:dataset}


\subsection{Simultaneous EEG-fMRI datasets}\label{subsection:dataset_description}
\vskip -0.15cm

\noindent The contributions of this work are assessed against two distinct neuroimaging datasets with simultaneous EEG-fMRI recordings: i) NODDI dataset with recordings conducted under resting states; and ii) Oddball dataset with stimuli-based recordings. These contrasting settings offer the possibility to acquire a comprehensive understanding on the ability to synthesize fMRI from EEG under different protocols. 
\vskip 0.2cm

\noindent\textbf{NODDI Dataset.} 
NODDI dataset \citep{dataset_noddi_1, dataset_noddi_2} contains $17$ individuals ($11$ males, $6$ females) with average age $32.84 \pm 8.13$ years. $10$ out of the $17$ individuals are considered, due to corrupted views. Simultaneous EEG-fMRI recordings of resting state with eyes open (fixating a point) were acquired. Subjects were told to stay still on a vaccum cushion during scanning. The fMRI imaging acquisition was done based on a T2-weighted gradient-echo EPI sequence with: $300$ volumes, TR of $2160$ milliseconds (ms), TE of $30$ ms, $30$ slices with $3.0$ millimeters (mm) (1 mm gap), voxel size of $3.3\times3.3\times4.0$ mm and a field of view of $210\times210\times120$ mm. The EEG imaging was acquired during the MRI scan with a $64$-channel-MR-compatible electrode cap at $1000$ Hz. The electrodes were setup according to the modified combinatorial nomenclature, referenced to the FCz electrode. An electrocardiogram (ECG) was recorded, and the EEG and MR scanner clocks were synchronised. 
The dataset is available for download by its original source at \url{https://osf.io/94c5t/}.
\vskip 0.2cm

\noindent\textbf{Auditory and visual Oddball Dataset.} 
The Oddball simultaneous EEG-fMRI dataset \citep{walz2013simultaneous02dataset,walz2014simultaneous02dataset,conroy2013fast02dataset} contains visual and auditory stimuli based recordings, each $340$ seconds long. In total, there are $17$ individuals out of which only $14$ were considered, due to corrupted views. The acquisition was made with a 3T Philips Achieva MR Scanner with: single channel send and receive head coil, EPI sequence, $170$ TRs per run with a TR of $2000$ ms and $25$ ms TE, $170$ TRs per run with a $3\times3\times4$ mm voxel size and $32$ slices with no slice gap. For a more detailed description of the dataset please refer to \citep{walz2013simultaneous02dataset}. The dataset is available for download in its original source at \url{https://legacy.openfmri.org/dataset/ds000116/}.

\subsection{Pairing EEG and fMRI: data setup}\label{approach:dataset_setup}

\vskip -0.15cm

\noindent EEG and BOLD data preprocessing was performed in accordance with \citet{dataset_noddi_1} and \citet{walz2013simultaneous02dataset} for the NODDI and Oddball datasets. 
In addition to the original preprocessing and \citet{lewis2005logarithmic} principles based on the observation that the distribution of fMRI values on the data at hand follow a lognormal distribution, we decided to log scale the fMRI values as well. The fMRI signal was downsampled (using the nilearn python library \citep{nilearn}) by a factor of $3$ due to its fine resolution (i.e. number of voxels). The Short-Time Fourier Transform (STFT) was computed on the EEG signal, due to the variation of frequency intensities being correlated with the BOLD signal \citep{portnova2018correlation}. The STFT was taken with a window of $2$ seconds and the frequency resolution placed according to the frequency sampling of the corresponding dataset.

According to \citet{liao2002593}, it is estimated that the neuronal activity is reflected in the BOLD signal with a delay of $s\approx[5.4,6]$ seconds. Pairs between EEG and BOLD should have a shift of $s$ seconds, such that at time $t_{EEG}$ then the corresponding BOLD pair starts at time $t_{BOLD}(= t_{EEG} + s)$. In addition, there is the need to specify a time window big enough to adequately decode the lower frequency bands from the EEG signal. The interval can further impact the number of features at input and output, making the problem much more difficult. This balance is extremely important as the network should not be forced to learn these properties. 

The datasets described in this section contain EEG and fMRI recordings lengthy enough to be divided into partitions of $25.2$ seconds each. In order for the bold shift $s=5.4$ to be emulated, both STFT EEG and BOLD signals were resampled to $1.8$ seconds using the \textsf{scipy} Python library \citep{scipy}.

\section{Proposed EEG to fMRI approach}\label{section:approach}

\noindent This section introduces the proposed approaches for EEG to fMRI synthesis (Section \ref{approach:models}), their hyperparameterization (Section \ref{approach:hyperparameter_tuning}) and evaluation (Section \ref{approach:evaluation}).

\subsection{Proposed models}\label{approach:models}
\vskip -0.15cm

Pairwise learning \cite{mencia2008pairwise} is considered within the proposed approaches, whereby positive and negative pairs of EEG-fMRI recordings are fed as input to guide the learning. 
The positive pairs correspond to the EEG and BOLD starting at $t_{EEG}$ and $t_{BOLD}$, respectively. Each positive pair of EEG and BOLD belong to the same individual and the same recording session. In contrast, the negative pairs are all the combinations of EEG and BOLD intances, that verify the following conditions: $t_{EEG} \neq t_{BOLD} + s$ and the individual corresponding to the EEG instance is different from the individual corresponding to the BOLD instance.

The proposed models have traits based on the AE, VAE and $\beta$-VAE, where instead of indirectly optimizing the parameters of a distribution (as it is the case for the VAE and $\beta$-VAE) the parameters being optimized are only the learnable network parameters (e.g. weights, biases, etc). In particular, the extended version of $\beta$-VAE considers reconstruction loss at the output level (similarly to the VAE and AE). This variant is subsequently reintroduced in the form of a linear combination with a distance loss (e.g. Contrastive Loss \citep{contrastive2005chopra}) loss at the midlayer level (as done in $\beta$-VAE). 

\begin{figure}[ht]
    \centering
    \includegraphics[width=0.48\textwidth]{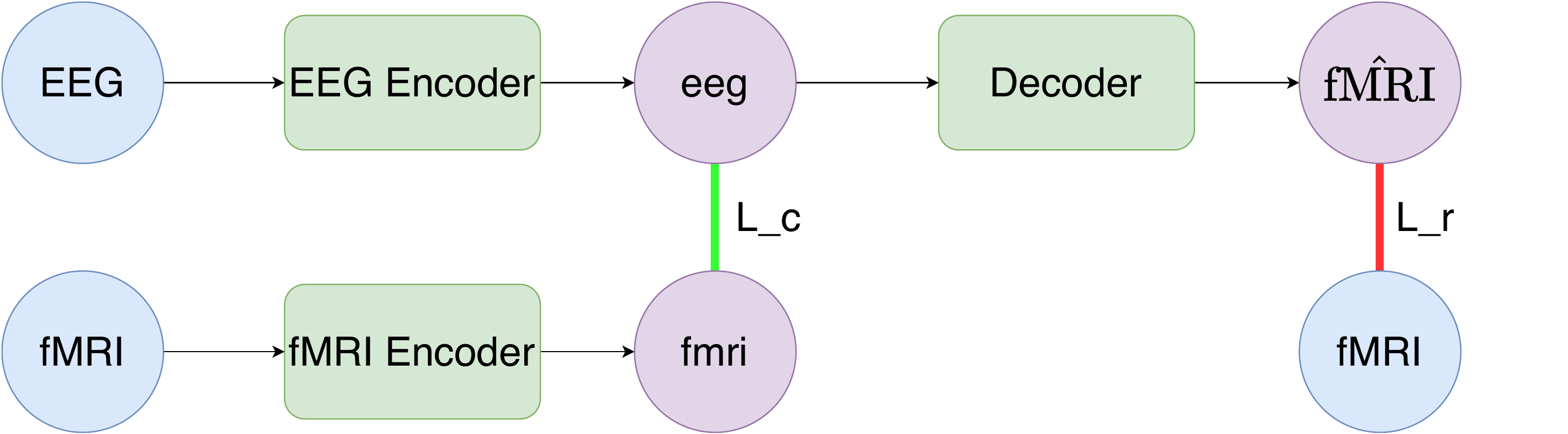}
        \vskip -0.1cm
    \caption{\small Pipeline of EEG to fMRI synthesis.}
        \vskip -0.2cm
    \label{fig:multi_modal_model}
\end{figure}

Figure \ref{fig:multi_modal_model} depicts the proposed EEG-to-fMRI synthesis pipeline taking into account positive and negative imaging pairs. In accordance with pairwise learning principles, fMRI Encoder is not consider at testing time. The number of layers, $N_L$, along the pipeline components is placed by Neural Architecture Search (NAS) \cite{elsken2018neural}. 
The EEG Encoder and the fMRI Encoder are both trained with the same loss, which varies depending on the training procedure. The Decoder is trained with a loss drawn from its output. 

Four classes of networks are proposed: \textbf{Linear Combination} (Section \ref{subsubsection:linear_combination_training}), \textbf{AE Baseline} (Section \ref{subsubsection:ae_training}), \textbf{Adversarial} (Section \ref{subsubsection:adversarial_training}) and \textbf{Top-}$\mathbf{k}$ \textbf{Ranking} (Section \ref{subsubsection:top_k_training}). Section \ref{subsubsection:temporal_regularization} introduces a technique that is capable of capturing temporal patterns at the encoder level, which is used by all the architectures.

\subsubsection{Linear Combination (LCOMB)}\label{subsubsection:linear_combination_training}

The reconstruction loss, $L_r$, is represented by the \textbf{Euclidean Per Volume}, $L_{EPV}(\text{fMRI}, \hat{\text{fMRI}})$,

$$
\small  \frac{ \sum_{i=0}^{N_{volumes}} {\frac{ \sqrt{ \sum_{v=0}^{N_{voxels}} (\text{fMRI}_{i,v} - \hat{\text{fMRI}}_{i,v}) ^2} } {N_{voxels}} } } {N_{volumes}} .
$$

By minimizing this loss function, one converges to an optimal synthesized representation of an fMRI signal from the paired EEG signal. The choice of the Euclidean Per Volume Loss over the Mean Absolute Error Loss was based on the first having lower magnitude values, which may have an impact in the gradients computation. 

In addition, to the $L_r$ being introduced at the output level, it is also reintroduced to the EEG Encoder and BOLD Encoder in a linear combination, $L_e$, with a Contrastive Loss, $L_c(W,Y,\text{EEG},\text{fMRI})$,

\vskip -0.15cm
$$
    Y{D_W}^2 + (1-Y) \ {max(0, m - D_W)^2} ,
$$
\vskip -0.4cm

$$
    L_e = \theta L_c + (1-\theta)L_r .
$$

Regarding $L_c$, (EEG,fMRI) is the input pair, $Y$=1 if $EEG$ and $fMRI$ are positive pairs and $0$ otherwise, $D_W$ the distance between the predicted values of $eeg$ and $fmri$, and $m$ is the margin value of separation. 

The Contrastive Loss function forces the neighbors to be pulled together and non-neighbors to be pushed apart. This loss uses a distance metric. The Mean Absolute Error is the chosen metric.

Encoders take into account not only the approximation of the $eeg$ and $fmri$ signals (by mapping the encoder outputs, $eeg$ and $fmri$ signals get closer in space for positive pairs), but also maintain the reconstruction properties of the signal when performing the mapping. Under this premise, Encoders have a loss, $L_e$, that is a linear combination (set by $\theta$) of $L_r$ and $L_c$.

\subsubsection{AE Baseline (AE)}\label{subsubsection:ae_training}

\begin{figure}[ht]
    \centering
    \includegraphics[width=0.48\textwidth]{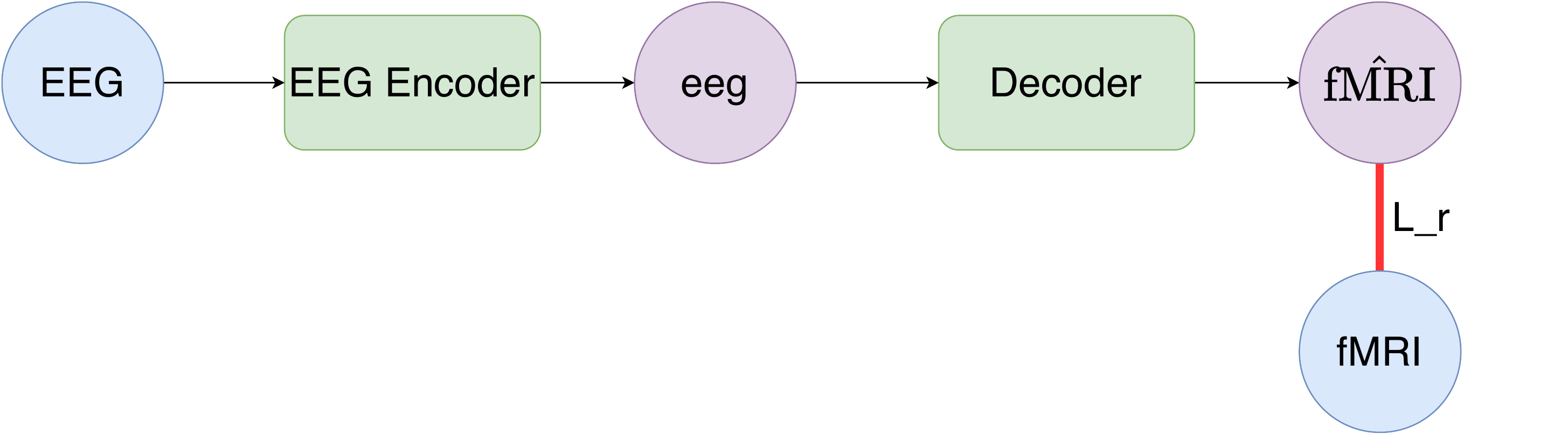}
        \vskip -0.2cm
    \caption{Auto-Encoder architecture.}
        \vskip -0.2cm
    \label{fig:auto_encoder}
\end{figure}

An AE model incorporating the architecture described in Figure \ref{fig:auto_encoder} is developed to be used as a baseline. This architecture is the one used in a test phase to perform the transcription from the EEG signal to the synthesized fMRI signal. The AE is treated as a baseline and is included in the results (see Section \ref{section:results}).

\subsubsection{Adversarial (GAN and WGAN)}\label{subsubsection:adversarial_training}

As discussed in Section \ref{section:related_work}, GANs have shown to be useful in cross-modalilty image synthesis, therefore this work also considers this type of deep learning approach for the synthesis of two functional neuroimaging modalities.

Although the $L_r$ loss forces the model to learn the spatial properties of the original signal, this may not be enough to make the signal as close as possible to the original. An adversarial learning process introduces penalties given by a Discriminator and Generator components. If the Discriminator recognizes instances synthesized by the Generator, then a penalization is given to the Generator. On the other hand, if the Discriminator does not recognize those synthesized instances, a penalization is given to Discriminator itself. We consider two variations of this type of learning: the Minmax Entropy Loss (also known as Vanilla \textbf{GAN}),
\vskip -0.3cm

$$
    \mathbb{E}_{x \sim p_{data}(x)}[log(D(X))] + \mathbb{E}_{x \sim p_{z}(z)}[log(1-D(G(z)))],
$$

\noindent and the Earth Mover Distance Loss (also known as \textbf{WGAN}),

\vskip -0.3cm

$$
    \mathbb{E}_{x \sim p_{data}(x)}[D(X)] + \mathbb{E}_{x \sim p_{z}(z)}[1-D(G(z))],
$$

\noindent where $x\sim p_{data}(x)$ is an instance taken from the real instances and $x\sim p_z(z)$ is a sample taken from a distribution, subsequently decoded by $G$ to $G(z)$.

\subsubsection{Top-k Ranking}\label{subsubsection:top_k_training}

We found it pertinent to implement another baseline inspired on information retrieval top-$k$ techniques. For that, this next variant concentrates on yet another variation at the encoding level, $eeg$. Instead of decoding directly the $eeg$ activations, a linear combination of top-$k$ $eeg$s is given to the Decoder. This linear combination is a normalized vector of correlation values from the most correlated $eeg$ instances. The EEG and fMRI Encoders are trained for a fixed number of epochs. Once this training session is over, each instance in the training set is compared to all the others, producing a rank of $eeg$ instances for each $eeg$ instance. Following, the top-$k$ $eeg$ instances are selected and a linear combination of these instances is computed,
\vskip -0.5cm

$$
    top\_k\_eeg = \sum_{r=0}^{k} {corr({eeg}_r, eeg) \times {eeg}_r} .
$$
\vskip -0.15cm

The Decoder then begins its training session with the inputs being the set of instances from the linear combinations and the targets the $fMRI$ associated with each $EEG$.

\subsubsection{Temporal Encoding}\label{subsubsection:temporal_regularization}

\begin{figure}[ht]
    \centering
    \includegraphics[width=0.5\textwidth]{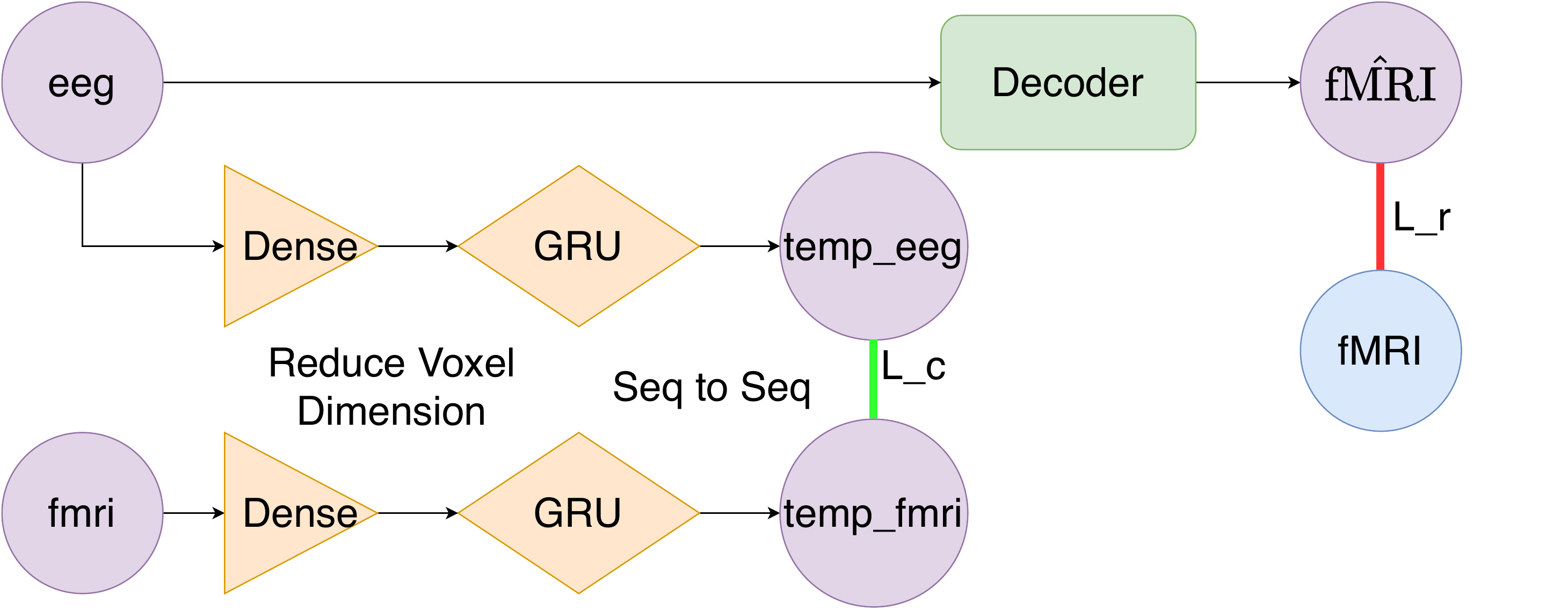}
    \caption{\small Temporal Encoding integration in the architecture shown in Figure \ref{fig:multi_modal_model}. At the $Decoder$ level, the convolutionally encoded $eeg$ is fed as input and the $L_r$ is computed against the ground truth $fMRI$. At the $Encoder$, the $eeg$ and $fmri$ go through a fully connected layer that squeezes the Voxel dimension to $1$, followed by a seq-to-seq Gated Recurrent Unit (GRU) layer that outputs $temp\_eeg$ and $temp\_fmri$, respectively. The pipeline from $eeg$ and $fmri$ are independent, i.e. do not share weights. $L_c$ is computed from the $temp\_eeg$ and $temp\_fmri$ temporal encoded activations.}
    \vskip -0.2cm
    \label{fig:temporal_encoding}
\end{figure}

Since, both modalities (EEG and fMRI) are functional neuroimaging techniques (i.e. contain temporal properties), there is a need for operations capable of capturing such properties. Most of the variants introduced so far, share a similar architecture to the one shown in Figure \ref{fig:multi_modal_model}, only using Convolutional and Convolutional Transposed layers (\citet{bai2018empirical} perform a thorough analysis on the performance of convolutional based networks against recurrent based networks, concluding that convolutions are preferred demonstrating effective memory properties). And although, these convolutions are performed on multiple dimensions (including the time dimension), the combination of these opreations along with recurrent layers is favorable \cite{quang2016danq,lopez2017network,zuo2017combining,mousavi2019cred}. As such, in order to answer to the main question of this work, a recurrent component that provides time dependent encodings is introduced. Figure \ref{fig:temporal_encoding} shows the setting used to incorporate the rationale explained.

\subsection{Hyperparameter Tuning}\label{approach:hyperparameter_tuning}
\vskip -0.15cm

\noindent Each of the variants described in Section \ref{approach:models} has different hyperparameters that are optimized. The tuning is done according to the performance of hyperparameters in a validation set. The hyperparameters that are common to all variants are: learning rate, weight regularization (L1 normalization) and batch size. In addition to those, \textbf{Linear Combination} needs the loss coefficient parameter, $\theta$, to be optimized as well. On the other hand, the $k$ value from the \textbf{Top-k Ranking} does not impact the performance and was fixed at $k=5$. 
With this, as the \textbf{Linear Combination} is the procedure that has more hyperparameters to be tuned and containing all the hyperparameters from the other variants, it is chosen to be subjected to a NAS \cite{elsken2018neural}. The Bayesian Optimization (BO) Algorithm is integrated in the search algorithm, therefore the hyperparameters are tuned along with the architecture. The search is done with $100$ BO iterations for each depth of the network, stopping when there is no improvement (on the validation set) at a certain deptth $d$ against the optimal hyperparameters discovered at $d-1$. The optimal hyperparameters given are discovered at depth $d-1$. 

The range of hyperparameters explored by the BO were ($p\_layer$ and $n\_layer$ represent the shapes of the previous and next layers, respectively): learning rate $\in [1\mathrm{e}{-14}, 1\mathrm{e}{-3}] \in \mathbb{R}$; L1 EEG Encoder regularization $\in [1\mathrm{e}{-5}, 1\mathrm{e}{-1}] \in \mathbb{R}$; L1 BOLD Encoder regularization $\in [1\mathrm{e}{-5}, 1\mathrm{e}{-1}] \in \mathbb{R}$; L1 Decoder regularization $\in [1\mathrm{e}{-5}, 1\mathrm{e}{-1}] \in \mathbb{R}$; loss coefficient, $\theta \in [0, 1] \in \mathbb{R}$; batch size $\in \{2, 4, 8, 16, 32, 64, 128\} \in \mathbb{N}$; EEG Encoder layer shape $\in [p\_layer, n\_layer] \in \mathbb{N}$; BOLD Encoder layer shape $\in [p\_layer, n\_layer] \in \mathbb{N}$; Decoder layer shape $\in [p\_layer, n\_layer] \in \mathbb{N}$. Dropout Layers \cite{dropout} follow after each added layer with a probability of dropping connections $p=0.5$.

\subsection{Evaluation Metrics}\label{approach:evaluation}
\vskip -0.15cm

To address the quality of the synthesized fMRI signals different metrics, exploring both temporal (BOLD) and spatial (fMRI) resolutions of the synthesized signals, are computed in addition to the Loss being minimized ($L_{EPV}$). The metrics computed are: \textbf{Log-Cosine Flattened Voxels (LCFV)}, \textbf{Cosine Flattened Voxels (CFV)}, \textbf{Euclidean Mean Voxels (EMV)}, \textbf{Euclidean Per Volume (EPV)}, \textbf{Mean Absolute Error (MAE)} and \textbf{Kullback–Leibler divergence (KL)}. \textbf{LCFV} computes the Log-Cosine of a flattened time series from all the voxels, evaluating the temporal resolution,
    \vskip -0.4cm

$$
    log(1-cosine(flatten(\text{BOLD}), flatten(\hat{\text{BOLD}}))).
$$

\textbf{CFV} computes the Cosine of a flattened time series from all the voxels, evaluating the temporal resolution,
    \vskip -0.4cm

$$
    cosine(flatten(\text{BOLD}), flatten(\hat{\text{BOLD}})).
$$

\textbf{EMV} computes the Mean of the Euclidean Distance of all the voxels, evaluating the temporal resolution,
    \vskip -0.1cm

$$
    \frac{ \sum_{i=0}^{N_{voxels}} {euclidean(\text{BOLD}_i, \hat{\text{BOLD}}_i)} }{N_{voxels}}.    
$$

\textbf{EPV} computes the Mean of the Euclidean of fMRI Volumes, evaluating the spatial resolution,
    \vskip -0.2cm

$$
    EPV = \frac{ \sum_{i=0}^{N_{volumes}} { \frac{ \sum_{v=0}^{N_{voxels}} \sqrt{(\text{fMRI}_{i,v} - \hat{\text{fMRI}}_{i,v})^2} } {N_{voxels}} }} {N_{volumes}}.
$$

\textbf{MAE} computes the Mean Absolute Error of fMRI Volumes, evaluating the spatial resolution,

$$
    \frac{ \sum_{i=0}^{N_{volumes}} { \frac{ \sum_{v=0}^{N_{voxels}} |\text{fMRI}_{i,v} - \hat{\text{fMRI}}_{i,v}| } {N_{voxels}} }} {N_{volumes}}.
$$

\textbf{KL} computes the Kullback–Leibler Divergence of the fMRI Volumes, evaluating the spatial resolution,
    \vskip -0.1cm

$$
    \frac{ \sum_{i=0}^{N_{volumes}} {KL(\text{fMRI}_i, \hat{\text{fMRI}}_i) }}{N_{volumes}}.
$$

\section{Results}\label{section:results}

\begin{figure*}[ht!]
    \centering
    \begin{subfigure}[b]{0.13\textwidth}
        \centering
        \includegraphics[width=\textwidth]{figures/fmri_real_1_13.png}
        \caption{Real fMRI}
        \label{fig:real_fmri_1}
    \end{subfigure}
    \hfill
    \begin{subfigure}[b]{0.13\textwidth}
        \centering
        \includegraphics[width=\textwidth]{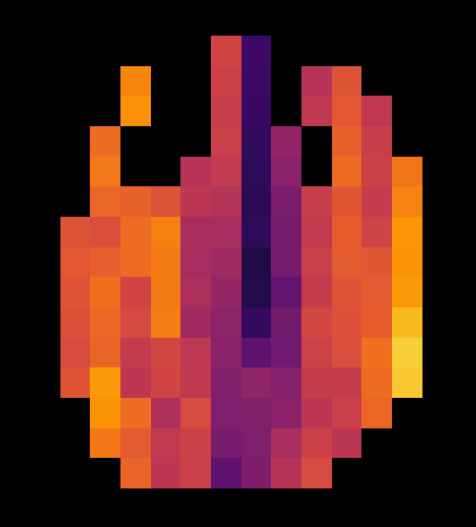}
        \caption{AE}
        \label{fig:ae_qual_results_1}
    \end{subfigure}
    \hfill
    \begin{subfigure}[b]{0.13\textwidth}
        \centering
        \includegraphics[width=\textwidth]{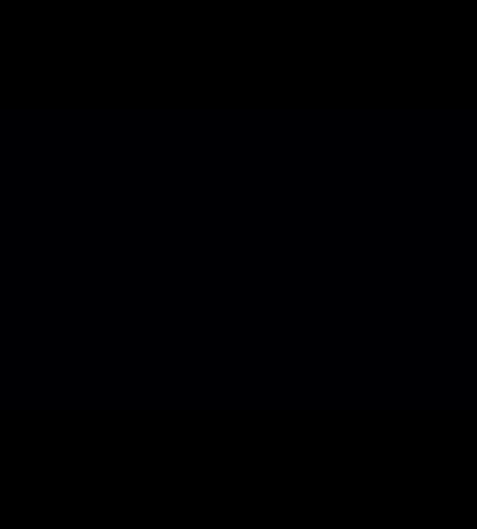}
        \caption{GAN}
        \label{fig:vgan_qual_results_1}
    \end{subfigure}
    \hfill
    \begin{subfigure}[b]{0.13\textwidth}
        \centering
        \includegraphics[width=\textwidth]{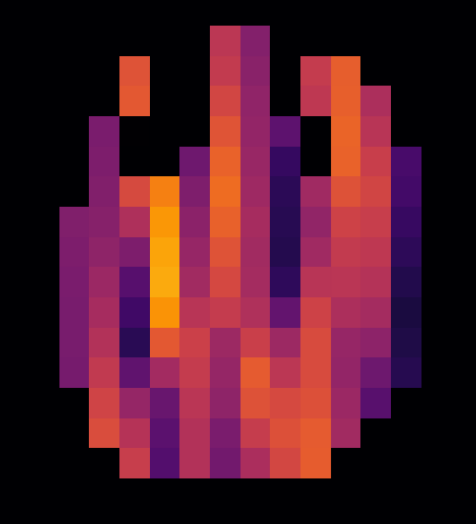}
        \caption{WGAN}
        \label{fig:gan_qual_results_1}
    \end{subfigure}
    \hfill
    \begin{subfigure}[b]{0.13\textwidth}
        \centering
        \includegraphics[width=\textwidth]{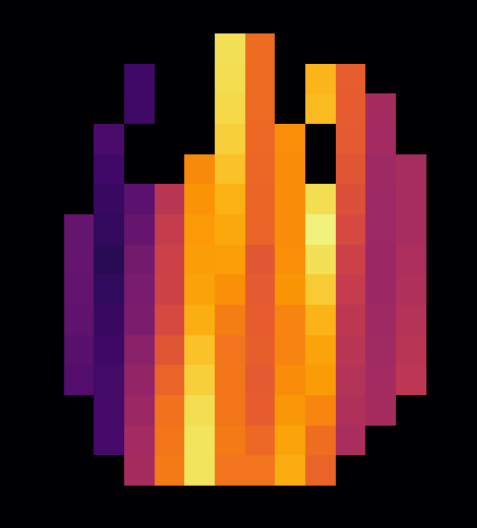}
        \caption{LCOMB}
        \label{fig:lcomb_qual_results_1}
    \end{subfigure}
    \hfill
    \begin{subfigure}[b]{0.13\textwidth}
        \centering
        \includegraphics[width=\textwidth]{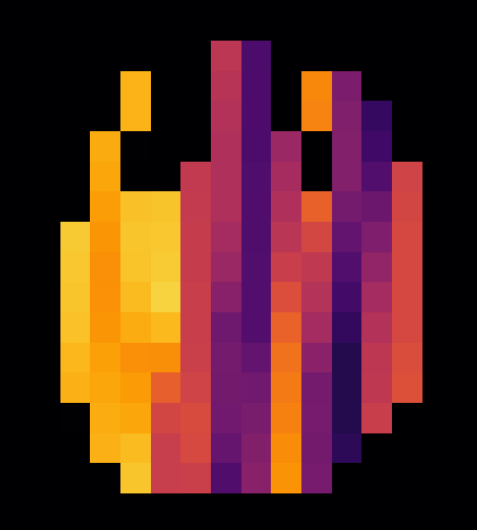}
        \caption{TOP-5}
        \label{fig:top5_qual_results_1}
    \end{subfigure}
    \vskip -0.2cm
    \caption{\small Qualitative results on NODDI data, corresponding to the second individual in the test set (timestep at $26$ seconds).}
    \vskip -0.1cm
    \label{fig:qual_results_01}
\end{figure*}

\begin{table*}[ht!]
\footnotesize
    \centering
    \setlength{\tabcolsep}{1.5pt}
    \begin{tabular}{l l l l l l}
        \hline
         & AE      & GAN     & WGAN      & LCOMB  & TOP-5\\
        \hline
        (i)  & $\mathbf{-0.252\pm0.324}$  & $\mathbf{-0.269\pm0.349}$ & $\mathbf{-0.267\pm0.345}$ & $-0.071\pm0.051$ & $\mathbf{-0.266\pm0.343}$ \\
        (ii)  & $\mathbf{0.193\pm0.165}$ & $\mathbf{0.202\pm0.169}$ & $\mathbf{0.201\pm0.168}$ & $0.067\pm0.047$ & $\mathbf{0.201\pm0.168}$\\
        (iii)  & $\mathbf{83.2\pm34.4}$ & $131\pm40.7$ & $102\pm38.7$ & $\mathbf{87.7\pm32.8}$ & $111\pm39.0$\\
        \hline
        (iv)  & $\mathbf{22.0\pm8.07}$ & $34.7\pm9.99$ & $27.1\pm9.3$ & $\mathbf{23.2\pm7.57}$ & $29.4\pm9.44$\\
        (v)  & $\mathbf{505\pm241}$ & $888\pm300$ & $657\pm279$ & $\mathbf{538\pm225}$ & $725\pm282$\\
        \hline
        (vi) & $0.475\pm0.204$ & $-0.131\pm0.025$ & $\mathbf{0.003\pm0.044}$ & $1.611\pm1.313$ & $\mathbf{-0.047\pm0.040}$\\
        \hline
    \end{tabular}
    \caption{Quantitative results on the NODDI Dataset.}
    \label{table:test_results_01}
\end{table*}

\begin{table*}[ht!]
\footnotesize
    \centering
    \setlength{\tabcolsep}{1.5pt}
    \begin{tabular}{l l l l l l}
        \hline
         & AE      & GAN     & WGAN      & LCOMB  & TOP-5\\
        \hline
        (i) & $-0.030\pm0.034$ & $-0.030\pm0.034$ & $\mathbf{-0.031\pm0.034}$ & $\mathbf{-0.031\pm0.034}$ & $\mathbf{-0.031\pm0.034}$ \\
        (ii) & $0.029\pm0.032$ & $\mathbf{0.030\pm0.032}$ & $\mathbf{0.030\pm0.032}$ & $\mathbf{0.030\pm0.032}$ & $\mathbf{0.030\pm0.032}$ \\ 
        (iii)  & $\mathbf{111\pm5.11}$ & $156\pm5.54$ & $139\pm5.52$ & $\mathbf{111\pm5.10}$ & $\mathbf{112\pm5.24}$\\
        \hline
        (iv)  & $\mathbf{29.5\pm1.32}$ & $41.7\pm1.45$ & $37.0\pm1.44$ & $\mathbf{29.5\pm1.32}$ & $\mathbf{29.9\pm1.36}$\\
        (v)  & $\mathbf{701\pm31.4}$ & $1024\pm36.7$ & $892\pm35.4$ & $\mathbf{701\pm31.3}$ & $\mathbf{710\pm32.3}$\\
        \hline
        (vi) & $1.717\pm0.267$ & $-0.089\pm0.004$ & $\mathbf{-0.001\pm0.005}$ & $5.023\pm0.252$ & $0.483\pm0.030$\\
        \hline
    \end{tabular}
    \caption{Quantitative results on Auditory and Visual Oddball Dataset.}
    \label{table:test_results_02}
\end{table*}

\noindent The qualitative results are presented in Figure \ref{fig:qual_results_01}.
The quantitative results gathered from the models for the test set ($2$ individuals and $4$ individuals for the NODDI Dataset and Auditory and Visual Oddball Dataset, respectively) are presented in Tables \ref{table:test_results_01} and \ref{table:test_results_02}. Each row in this table corresponds to the metrics described in Section \ref{approach:evaluation}, using the following numeration: (i) LCFV, (ii) CFV, (iii) EMV, (iv) EPV, (v) MAE and (vi) KL.

As for the results gathered on the NODDI Dataset present in Table \ref{table:test_results_01}, AE had the best results in terms of the metrics evaluated (quantitative results), at the naked eye it seems to have good qualitative results (see Figure \ref{fig:ae_qual_results_1}), but on the other hand its KL metric evaluation was poor compared to the others. GAN and WGAN did not target the loss common to the other models ($L_r$), which possibly explains their inferiority under spatial metrics (iv) and (v). WGAN showed its capacity to synthesize fMRI signal taking into account their mean and variability as it is shown by (vi) KL, which also impacted the quality of the synthesized signal in Figure \ref{fig:gan_qual_results_1}. In regard to pattern based metrics (i) LCFV and (ii) CFV, although there is no model that has a clear superiority, LCOMB had an inferior performance. In contrast, given an Euclidean evaluation along the time axis (pattern based) the LCOMB model performs among the best along with AE. Spatial based metrics showed AE and LCOMB are preferable, this is justifiable with the loss targeted being a spatial loss, described by (iv) EPV. Regarding the synthesis quality of LCOMB, Figure \ref{fig:lcomb_qual_results_1} shows a bandy pattern and a poor distribution, being concordant with the poor KL.

As for the results gathered on the Auditory and Visual Oddball Dataset present in Table \ref{table:test_results_02}, AE, LCOMB and TOP-5 had the best results, in terms of spatial resolution, given by the (iv) and (v) metrics. As for pattern based metrics, there was no clear difference when looking at the (i) and (ii) metrics. On the other hand, GAN and WGAN underperformed according to (iii). In terms of the values distribution evaluated by KL (vi), WGAN had once again the closest value to $0.0$, i.e. it had the best performance in this aspect. Regarding the qualitative results, at the naked eye, the quality of the synthesized signals was very poor for this dataset 
, with WGAN having a small finer superiority.

Qualitative results of GAN and TOP-5 synthesis were extremely poor with nothing synthesized for most cases, as these methods produced non defined values due to exploding gradients (clipping the loss seemed to have no effect). GAN loss computes a logarithm having a bigger magnitude than the earth mover distance loss of WGAN. As for the TOP-5, it seems a linear combination of $eeg$s at the encoder level does no resemble a good representation at that level. Overall, given that WGAN reaches the other models by the metrics evaluated and most importantly outperforms others in the (vi) metric on both datasets, which is shown in its qualitative results in Figure \ref{fig:gan_qual_results_1}
, it is seen as the best fit candidate for this task, among the ones covered in this manuscript. Further, the need for more data was shown, as a simple model such as AE had good quantitative and qualitative results. This is due to its lower number of learnable parameters. Nonetheless, it can still be claimed that WGAN (while having a high number parameters, for amount of data available) is more suitable for this task.

\section{Conclusion}\label{section:conclusion}

This manuscript provides compelling empirical evidence for the feasibility of relating haemodynamics and electrophysiology in the human brain as given by the study of fMRI data synthesis from EEG data. To this end, we proposed approaches grounded on state-of-the-art principles of neural processing, including pairwise and adversarial learning. Of particular interest, the Contrastive Loss \citep{hadsell2006dimensionality} trait for separating neighbours at the encoding level is useful for the targeted synthesis task. 
The gathered results further motivate the relevance of upcoming contributions to the targeted synthesis task, offering solid baselines of performance. 

EEG to fMRI synthesis task is expected to have major advances in the following decade, with broad applications in fields such as health, computer vision and neuroscience. Research on these tasks offers new ways of enriching brain imaging modalities, gaining further insights into the brain, and promoting long-lasting and less-expensive monitoring protocols.


\vskip 0.2cm 
\noindent\textbf{Future work}. The reverse transformation, fMRI-to-EEG, is also of high relevance to the community. Complementary cohort studies with simultaneous EEG and fMRI monitoring are being undertaken \cite{abreu2018eeg}, untapping new possibilities. Alternative approaches combining alternative principles from signal processing and time series data analysis are also expected. 
Finally, the role of emerging state-of-the-art neural processing techniques, such as Neural ODEs \citep{chen2018neural}, to this specific synthesis task is still unexplored.

\scriptsize
\bibliographystyle{plainnat}
\bibliography{egbib}

\begin{thebibliography}{58}
\providecommand{\natexlab}[1]{#1}
\providecommand{\url}[1]{\texttt{#1}}
\expandafter\ifx\csname urlstyle\endcsname\relax
  \providecommand{\doi}[1]{doi: #1}\else
  \providecommand{\doi}{doi: \begingroup \urlstyle{rm}\Url}\fi

\bibitem[Abraham et~al.(2014)Abraham, Pedregosa, Eickenberg, Gervais, Mueller,
  Kossaifi, Gramfort, Thirion, and Varoquaux]{nilearn}
Alexandre Abraham, Fabian Pedregosa, Michael Eickenberg, Philippe Gervais,
  Andreas Mueller, Jean Kossaifi, Alexandre Gramfort, Bertrand Thirion, and
  Gael Varoquaux.
\newblock Machine learning for neuroimaging with scikit-learn.
\newblock \emph{Frontiers in Neuroinformatics}, 8:\penalty0 14, 2014.
\newblock ISSN 1662-5196.
\newblock \doi{10.3389/fninf.2014.00014}.
\newblock URL
  \url{https://www.frontiersin.org/article/10.3389/fninf.2014.00014}.

\bibitem[Abramian and Eklund(2019)]{abramian2019generating}
David Abramian and Anders Eklund.
\newblock Generating fmri volumes from t1-weighted volumes using 3d cyclegan.
\newblock \emph{arXiv preprint arXiv:1907.08533}, 2019.

\bibitem[Abreu et~al.(2018)Abreu, Leal, and Figueiredo]{abreu2018eeg}
Rodolfo Abreu, Alberto Leal, and Patr{\'\i}cia Figueiredo.
\newblock Eeg-informed fmri: a review of data analysis methods.
\newblock \emph{Frontiers in human neuroscience}, 12:\penalty0 29, 2018.

\bibitem[Aggarwal and Gupta(2016)]{aggarwal2016accelerated}
Priya Aggarwal and Anubha Gupta.
\newblock Accelerated fmri reconstruction using matrix completion with sparse
  recovery via split bregman.
\newblock \emph{Neurocomputing}, 216:\penalty0 319--330, 2016.

\bibitem[Arjovsky et~al.(2017)Arjovsky, Chintala, and
  Bottou]{arjovsky2017wassersteing}
Mart{\'i}n Arjovsky, Soumith Chintala, and L{\'e}on Bottou.
\newblock Wasserstein gan.
\newblock \emph{ArXiv}, abs/1701.07875, 2017.

\bibitem[Bai et~al.(2018)Bai, Kolter, and Koltun]{bai2018empirical}
Shaojie Bai, J~Zico Kolter, and Vladlen Koltun.
\newblock An empirical evaluation of generic convolutional and recurrent
  networks for sequence modeling.
\newblock \emph{arXiv preprint arXiv:1803.01271}, 2018.

\bibitem[Ben-Cohen et~al.(2019)Ben-Cohen, Klang, Raskin, Soffer, Ben-Haim,
  Konen, Amitai, and Greenspan]{ben2019cross}
Avi Ben-Cohen, Eyal Klang, Stephen~P Raskin, Shelly Soffer, Simona Ben-Haim,
  Eli Konen, Michal~Marianne Amitai, and Hayit Greenspan.
\newblock Cross-modality synthesis from ct to pet using fcn and gan networks
  for improved automated lesion detection.
\newblock \emph{Engineering Applications of Artificial Intelligence},
  78:\penalty0 186--194, 2019.

\bibitem[Chang et~al.(2013)Chang, Liu, Chen, Liu, and Duyn]{chang2013eeg}
Catie Chang, Zhongming Liu, Michael~C Chen, Xiao Liu, and Jeff~H Duyn.
\newblock Eeg correlates of time-varying bold functional connectivity.
\newblock \emph{Neuroimage}, 72:\penalty0 227--236, 2013.

\bibitem[Chao et~al.(2018)Chao, Chang, Li, Wu, and Lee]{chao2018generating}
Gao-Yi Chao, Chun-Min Chang, Jeng-Lin Li, Ya-Tse Wu, and Chi-Chun Lee.
\newblock Generating fmri-enriched acoustic vectors using a cross-modality
  adversarial network for emotion recognition.
\newblock In \emph{Proceedings of the 20th ACM International Conference on
  Multimodal Interaction}, pages 55--62, 2018.

\bibitem[Chen et~al.(2018)Chen, Rubanova, Bettencourt, and
  Duvenaud]{chen2018neural}
Tian~Qi Chen, Yulia Rubanova, Jesse Bettencourt, and David~K Duvenaud.
\newblock Neural ordinary differential equations.
\newblock In \emph{Advances in neural information processing systems}, pages
  6571--6583, 2018.

\bibitem[Chen et~al.(2016)Chen, Kingma, Salimans, Duan, Dhariwal, Schulman,
  Sutskever, and Abbeel]{chen2016variational}
Xi~Chen, Diederik~P Kingma, Tim Salimans, Yan Duan, Prafulla Dhariwal, John
  Schulman, Ilya Sutskever, and Pieter Abbeel.
\newblock Variational lossy autoencoder.
\newblock \emph{arXiv preprint arXiv:1611.02731}, 2016.

\bibitem[{Chopra} et~al.(2005){Chopra}, {Hadsell}, and
  {LeCun}]{contrastive2005chopra}
S.~{Chopra}, R.~{Hadsell}, and Y.~{LeCun}.
\newblock Learning a similarity metric discriminatively, with application to
  face verification.
\newblock In \emph{2005 IEEE Computer Society Conference on Computer Vision and
  Pattern Recognition (CVPR'05)}, volume~1, pages 539--546 vol. 1, June 2005.
\newblock \doi{10.1109/CVPR.2005.202}.

\bibitem[Conroy et~al.(2013)Conroy, Walz, and Sajda]{conroy2013fast02dataset}
Bryan~R Conroy, Jennifer~M Walz, and Paul Sajda.
\newblock Fast bootstrapping and permutation testing for assessing
  reproducibility and interpretability of multivariate fmri decoding models.
\newblock \emph{PloS one}, 8\penalty0 (11), 2013.

\bibitem[Cury et~al.(2019)Cury, Maurel, Gribonval, and
  Barillot]{cury2019sparse}
Claire Cury, Pierre Maurel, R{\'e}mi Gribonval, and Christian Barillot.
\newblock A sparse eeg-informed fmri model for hybrid eeg-fmri neurofeedback
  prediction.
\newblock \emph{bioRxiv}, page 599589, 2019.

\bibitem[Deligianni et~al.(2014)Deligianni, Centeno, Carmichael, and
  Clayden]{dataset_noddi_1}
Fani Deligianni, Maria Centeno, David~W. Carmichael, and Jonathan~D. Clayden.
\newblock Relating resting-state fmri and eeg whole-brain connectomes across
  frequency bands.
\newblock \emph{Frontiers in Neuroscience}, 8:\penalty0 258, 2014.
\newblock ISSN 1662-453X.
\newblock \doi{10.3389/fnins.2014.00258}.
\newblock URL
  \url{https://www.frontiersin.org/article/10.3389/fnins.2014.00258}.

\bibitem[Deligianni et~al.(2016)Deligianni, Carmichael, Zhang, Clark, and
  Clayden]{dataset_noddi_2}
Fani Deligianni, David~W. Carmichael, Gary~H. Zhang, Chris~A. Clark, and
  Jonathan~D. Clayden.
\newblock Noddi and tensor-based microstructural indices as predictors of
  functional connectivity.
\newblock \emph{PLOS ONE}, 11\penalty0 (4):\penalty0 1--17, 04 2016.
\newblock \doi{10.1371/journal.pone.0153404}.
\newblock URL \url{https://doi.org/10.1371/journal.pone.0153404}.

\bibitem[Elsken et~al.(2018)Elsken, Metzen, and Hutter]{elsken2018neural}
Thomas Elsken, Jan~Hendrik Metzen, and Frank Hutter.
\newblock Neural architecture search: A survey.
\newblock \emph{arXiv preprint arXiv:1808.05377}, 2018.

\bibitem[Fowle and Binnie(2000)]{fowle2000uses}
Adrian~J Fowle and Colin~D Binnie.
\newblock Uses and abuses of the eeg in epilepsy.
\newblock \emph{Epilepsia}, 41:\penalty0 S10--S18, 2000.

\bibitem[Goodfellow et~al.(2014)Goodfellow, Pouget-Abadie, Mirza, Xu,
  Warde-Farley, Ozair, Courville, and Bengio]{goodfellow2014generativean}
Ian~J. Goodfellow, Jean Pouget-Abadie, Mehdi Mirza, Bing Xu, David
  Warde-Farley, Sherjil Ozair, Aaron~C. Courville, and Yoshua Bengio.
\newblock Generative adversarial nets.
\newblock In \emph{NIPS}, 2014.

\bibitem[Gramfort et~al.(2013)Gramfort, Luessi, Larson, Engemann, Strohmeier,
  Brodbeck, Goj, Jas, Brooks, Parkkonen, and Hämäläinen]{gramfort2013mne}
Alexandre Gramfort, Martin Luessi, Eric Larson, Denis Engemann, Daniel
  Strohmeier, Christian Brodbeck, Roman Goj, Mainak Jas, Teon Brooks, Lauri
  Parkkonen, and Matti Hämäläinen.
\newblock Meg and eeg data analysis with mne-python.
\newblock \emph{Frontiers in Neuroscience}, 7:\penalty0 267, 2013.
\newblock ISSN 1662-453X.
\newblock \doi{10.3389/fnins.2013.00267}.
\newblock URL
  \url{https://www.frontiersin.org/article/10.3389/fnins.2013.00267}.

\bibitem[Hadsell et~al.(2006)Hadsell, Chopra, and
  LeCun]{hadsell2006dimensionality}
Raia Hadsell, Sumit Chopra, and Yann LeCun.
\newblock Dimensionality reduction by learning an invariant mapping.
\newblock In \emph{2006 IEEE Computer Society Conference on Computer Vision and
  Pattern Recognition (CVPR'06)}, volume~2, pages 1735--1742. IEEE, 2006.

\bibitem[He et~al.(2019)He, Mo, Wang, Liu, Yang, and Cheng]{He_2019_CVPR}
Xiangyu He, Zitao Mo, Peisong Wang, Yang Liu, Mingyuan Yang, and Jian Cheng.
\newblock Ode-inspired network design for single image super-resolution.
\newblock In \emph{Proceedings of the IEEE/CVF Conference on Computer Vision
  and Pattern Recognition (CVPR)}, June 2019.

\bibitem[He et~al.(2018)He, Steines, Sommer, Gebhardt, Nagels, Sammer, Kircher,
  and Straube]{he2018spatialtemporaldo}
Yifei He, Miriam Steines, Jens Sommer, Helge Gebhardt, Arne Nagels, Gebhard
  Sammer, Tilo T.~J. Kircher, and Benjamin Straube.
\newblock Spatial–temporal dynamics of gesture–speech integration: a
  simultaneous eeg-fmri study.
\newblock \emph{Brain Structure and Function}, 223:\penalty0 3073--3089, 2018.

\bibitem[Higgins et~al.(2017)Higgins, Matthey, Pal, Burgess, Glorot, Botvinick,
  Mohamed, and Lerchner]{higgins2017betavae}
Irina Higgins, Lo{\"i}c Matthey, Arka Pal, Christopher Burgess, Xavier Glorot,
  Matthew~M Botvinick, Shakir Mohamed, and Alexander Lerchner.
\newblock beta-vae: Learning basic visual concepts with a constrained
  variational framework.
\newblock In \emph{ICLR}, 2017.

\bibitem[Hinton and Salakhutdinov(2006)]{hinton2006reducing}
Geoffrey~E Hinton and Ruslan~R Salakhutdinov.
\newblock Reducing the dimensionality of data with neural networks.
\newblock \emph{science}, 313\penalty0 (5786):\penalty0 504--507, 2006.

\bibitem[Hore and Ziou(2010)]{hore2010image}
Alain Hore and Djemel Ziou.
\newblock Image quality metrics: Psnr vs. ssim.
\newblock In \emph{2010 20th International Conference on Pattern Recognition},
  pages 2366--2369. IEEE, 2010.

\bibitem[Jiang et~al.(2020)Jiang, Li, Zhao, Xiao, Zhang, Sun, Yang, and
  Zhu]{jiang2020targeting}
Yihan Jiang, Zheng Li, Yang Zhao, Xiang Xiao, Wei Zhang, Peipei Sun, Yihong
  Yang, and Chaozhe Zhu.
\newblock Targeting brain functions from the scalp: Transcranial brain atlas
  based on large-scale fmri data synthesis.
\newblock \emph{NeuroImage}, 210:\penalty0 116550, 2020.

\bibitem[Jones et~al.(2001)Jones, Oliphant, Peterson, et~al.]{scipy}
Eric Jones, Travis Oliphant, Pearu Peterson, et~al.
\newblock {SciPy}: Open source scientific tools for {Python}, 2001.
\newblock URL \url{http://www.scipy.org/}.
\newblock [Online; accessed ].

\bibitem[Karras et~al.(2019)Karras, Laine, and Aila]{karrasla19}
Tero Karras, Samuli Laine, and Timo Aila.
\newblock A style-based generator architecture for generative adversarial
  networks.
\newblock In \emph{{IEEE} Conference on Computer Vision and Pattern
  Recognition, {CVPR} 2019, Long Beach, CA, USA, June 16-20, 2019}, pages
  4401--4410, 2019.
\newblock URL
  \url{http://openaccess.thecvf.com/content\_CVPR\_2019/html/Karras\_A\_Style-Based\_Generator\_Architecture\_for\_Generative\_Adversarial\_Networks\_CVPR\_2019\_paper.html}.

\bibitem[Kingma and Welling(2013)]{kingma2013autoencodingvb}
Diederik~P. Kingma and Max Welling.
\newblock Auto-encoding variational bayes.
\newblock \emph{CoRR}, abs/1312.6114, 2013.

\bibitem[Labounek et~al.(2019)Labounek, Bridwell, Marecek, Lamos, and
  Jan]{labounek2019eegsp}
Ren{\'e} Labounek, David~A. Bridwell, Radek Marecek, Martin Lamos, and Jir{\'i}
  Jan.
\newblock Eeg spatiospectral patterns and their link to fmri bold signal via
  variable hemodynamic response functions.
\newblock \emph{Journal of Neuroscience Methods}, 318:\penalty0 34--46, 2019.

\bibitem[Leite et~al.(2013)Leite, Leal, and Figueiredo]{leite2013transfer}
Marco Leite, Alberto Leal, and Patr{\'\i}cia Figueiredo.
\newblock Transfer function between eeg and bold signals of epileptic activity.
\newblock \emph{Frontiers in neurology}, 4:\penalty0 1, 2013.

\bibitem[Lewis et~al.(2005)Lewis, Jerde, Tzagarakis, Gourtzelidis,
  Georgopoulos, Tsekos, Amirikian, Kim, U{\u{g}}urbil, and
  Georgopoulos]{lewis2005logarithmic}
Scott~M Lewis, Trenton~A Jerde, Charidimos Tzagarakis, Pavlos Gourtzelidis,
  Maria-Alexandra Georgopoulos, Nikolaos Tsekos, Bagrat Amirikian, Seong-Gi
  Kim, K{\^a}mil U{\u{g}}urbil, and Apostolos~P Georgopoulos.
\newblock Logarithmic transformation for high-field bold fmri data.
\newblock \emph{Experimental brain research}, 165\penalty0 (4):\penalty0
  447--453, 2005.

\bibitem[Liao et~al.(2002)Liao, Worsley, Poline, Aston, Duncan, and
  Evans]{liao2002593}
C.H. Liao, K.J. Worsley, J.-B. Poline, J.A.D. Aston, G.H. Duncan, and A.C.
  Evans.
\newblock Estimating the delay of the fmri response.
\newblock \emph{NeuroImage}, 16\penalty0 (3, Part A):\penalty0 593 -- 606,
  2002.
\newblock ISSN 1053-8119.
\newblock \doi{https://doi.org/10.1006/nimg.2002.1096}.
\newblock URL
  \url{http://www.sciencedirect.com/science/article/pii/S1053811902910967}.

\bibitem[Lopez-Martin et~al.(2017)Lopez-Martin, Carro, Sanchez-Esguevillas, and
  Lloret]{lopez2017network}
Manuel Lopez-Martin, Belen Carro, Antonio Sanchez-Esguevillas, and Jaime
  Lloret.
\newblock Network traffic classifier with convolutional and recurrent neural
  networks for internet of things.
\newblock \emph{IEEE Access}, 5:\penalty0 18042--18050, 2017.

\bibitem[Menc{\'\i}a and Furnkranz(2008)]{mencia2008pairwise}
Eneldo~Loza Menc{\'\i}a and Johannes Furnkranz.
\newblock Pairwise learning of multilabel classifications with perceptrons.
\newblock In \emph{2008 IEEE International Joint Conference on Neural Networks
  (IEEE World Congress on Computational Intelligence)}, pages 2899--2906. IEEE,
  2008.

\bibitem[Mikulic(2019)]{mikulic_2019}
Matej Mikulic.
\newblock Mri units density by country 2017, Aug 2019.
\newblock URL
  \url{https://www.statista.com/statistics/282401/density-of-magnetic-resonance-imaging-units-by-country/}.

\bibitem[Mirza and Osindero(2014)]{mirza2014conditionalga}
Mehdi Mirza and Simon Osindero.
\newblock Conditional generative adversarial nets.
\newblock \emph{ArXiv}, abs/1411.1784, 2014.

\bibitem[Mosayebi and Hossein-Zadeh(2020)]{mosayebi2020correlated}
Raziyeh Mosayebi and Gholam-Ali Hossein-Zadeh.
\newblock Correlated coupled matrix tensor factorization method for
  simultaneous eeg-fmri data fusion.
\newblock \emph{Biomedical Signal Processing and Control}, 62:\penalty0 102071,
  2020.

\bibitem[Mousavi et~al.(2019)Mousavi, Zhu, Sheng, and Beroza]{mousavi2019cred}
S~Mostafa Mousavi, Weiqiang Zhu, Yixiao Sheng, and Gregory~C Beroza.
\newblock Cred: A deep residual network of convolutional and recurrent units
  for earthquake signal detection.
\newblock \emph{Scientific reports}, 9\penalty0 (1):\penalty0 1--14, 2019.

\bibitem[Murta et~al.(2015)Murta, Leite, Carmichael, Figueiredo, and
  Lemieux]{murta2015electrophysiologicalco}
Teresa Murta, Marco Leite, David~W. Carmichael, Patricia Figueiredo, and Louis
  Lemieux.
\newblock Electrophysiological correlates of the bold signal for eeg-informed
  fmri.
\newblock In \emph{Human brain mapping}, 2015.

\bibitem[Nie et~al.(2017)Nie, Trullo, Lian, Petitjean, Ruan, Wang, and
  Shen]{dong2017}
Dong Nie, Roger Trullo, Jun Lian, Caroline Petitjean, Su~Ruan, Qian Wang, and
  Dinggang Shen.
\newblock Medical image synthesis with context-aware generative adversarial
  networks.
\newblock In Maxime Descoteaux, Lena Maier-Hein, Alfred Franz, Pierre Jannin,
  D.~Louis Collins, and Simon Duchesne, editors, \emph{Medical Image Computing
  and Computer Assisted Intervention MICCAI 2017}, pages 417--425, Cham, 2017.
  Springer International Publishing.

\bibitem[Ogbole et~al.(2018)Ogbole, Adeyomoye, Badu-Peprah, Mensah, and
  Nzeh]{ogbole2018survey}
Godwin~Inalegwu Ogbole, Adekunle~Olakunle Adeyomoye, Augustina Badu-Peprah, Yaw
  Mensah, and Donald~Amasike Nzeh.
\newblock Survey of magnetic resonance imaging availability in west africa.
\newblock \emph{Pan African Medical Journal}, 30\penalty0 (1), 2018.

\bibitem[Portnova et~al.(2018)Portnova, Tetereva, Balaev, Atanov, Skiteva,
  Ushakov, Ivanitsky, and Martynova]{portnova2018correlation}
Galina~V Portnova, Alina Tetereva, Vladislav Balaev, Mikhail Atanov, Lyudmila
  Skiteva, Vadim Ushakov, Alexey Ivanitsky, and Olga Martynova.
\newblock Correlation of bold signal with linear and nonlinear patterns of eeg
  in resting state eeg-informed fmri.
\newblock \emph{Frontiers in human neuroscience}, 11:\penalty0 654, 2018.

\bibitem[Quang and Xie(2016)]{quang2016danq}
Daniel Quang and Xiaohui Xie.
\newblock Danq: a hybrid convolutional and recurrent deep neural network for
  quantifying the function of dna sequences.
\newblock \emph{Nucleic acids research}, 44\penalty0 (11):\penalty0 e107--e107,
  2016.

\bibitem[Reed et~al.(2016)Reed, Akata, Yan, Logeswaran, Schiele, and
  Lee]{reed2016generativeat}
Scott~E. Reed, Zeynep Akata, Xinchen Yan, Lajanugen Logeswaran, Bernt Schiele,
  and Honglak Lee.
\newblock Generative adversarial text to image synthesis.
\newblock \emph{ArXiv}, abs/1605.05396, 2016.

\bibitem[Rosa et~al.(2010)Rosa, Kilner, Blankenburg, Josephst, and
  Penny]{rosa2010estimating}
Maria~J Rosa, James Kilner, Felix Blankenburg, Oliver Josephst, and W~Penny.
\newblock Estimating the transfer function from neuronal activity to bold using
  simultaneous eeg-fmri.
\newblock \emph{Neuroimage}, 49\penalty0 (2):\penalty0 1496--1509, 2010.

\bibitem[Rumelhart et~al.(1986)Rumelhart, Hinton, and
  Williams]{rumelharthinton1986ae}
David~E. Rumelhart, Geoffrey~E. Hinton, and Ronald~J. Williams.
\newblock Learning internal representations by error propagation.
\newblock In David~E. Rumelhart and James~L. Mcclelland, editors,
  \emph{Parallel Distributed Processing: Explorations in the Microstructure of
  Cognition, {V}olume 1: {F}oundations}, pages 318--362. MIT Press, Cambridge,
  MA, 1986.

\bibitem[Srivastava et~al.(2014)Srivastava, Hinton, Krizhevsky, Sutskever, and
  Salakhutdinov]{dropout}
Nitish Srivastava, Geoffrey Hinton, Alex Krizhevsky, Ilya Sutskever, and Ruslan
  Salakhutdinov.
\newblock Dropout: A simple way to prevent neural networks from overfitting.
\newblock \emph{Journal of Machine Learning Research}, 15:\penalty0 1929--1958,
  2014.
\newblock URL \url{http://jmlr.org/papers/v15/srivastava14a.html}.

\bibitem[Turek et~al.(2017)Turek, Willke, Chen, and Ramadge]{turek2017semi}
Javier~S Turek, Theodore~L Willke, Po-Hsuan Chen, and Peter~J Ramadge.
\newblock A semi-supervised method for multi-subject fmri functional alignment.
\newblock In \emph{2017 IEEE International Conference on Acoustics, Speech and
  Signal Processing (ICASSP)}, pages 1098--1102. IEEE, 2017.

\bibitem[Walz et~al.(2013)Walz, Goldman, Carapezza, Muraskin, Brown, and
  Sajda]{walz2013simultaneous02dataset}
Jennifer~M Walz, Robin~I Goldman, Michael Carapezza, Jordan Muraskin, Truman~R
  Brown, and Paul Sajda.
\newblock Simultaneous eeg-fmri reveals temporal evolution of coupling between
  supramodal cortical attention networks and the brainstem.
\newblock \emph{Journal of Neuroscience}, 33\penalty0 (49):\penalty0
  19212--19222, 2013.

\bibitem[Walz et~al.(2014)Walz, Goldman, Carapezza, Muraskin, Brown, and
  Sajda]{walz2014simultaneous02dataset}
Jennifer~M Walz, Robin~I Goldman, Michael Carapezza, Jordan Muraskin, Truman~R
  Brown, and Paul Sajda.
\newblock Simultaneous eeg--fmri reveals a temporal cascade of task-related and
  default-mode activations during a simple target detection task.
\newblock \emph{Neuroimage}, 102:\penalty0 229--239, 2014.

\bibitem[Wei et~al.(2020)Wei, Jafarian, Zeidman, Litvak, Razi, Hu, and
  Friston]{wei2020bayesian}
Huilin Wei, Amirhossein Jafarian, Peter Zeidman, Vladimir Litvak, Adeel Razi,
  Dewen Hu, and Karl~J Friston.
\newblock Bayesian fusion and multimodal dcm for eeg and fmri.
\newblock \emph{NeuroImage}, 211:\penalty0 116595, 2020.

\bibitem[Wolterink et~al.(2017)Wolterink, Dinkla, Savenije, Seevinck, van~den
  Berg, and I{\v{s}}gum]{wolterink2017}
Jelmer~M. Wolterink, Anna~M. Dinkla, Mark H.~F. Savenije, Peter~R. Seevinck,
  Cornelis A.~T. van~den Berg, and Ivana I{\v{s}}gum.
\newblock Deep mr to ct synthesis using unpaired data.
\newblock In Sotirios~A. Tsaftaris, Ali Gooya, Alejandro~F. Frangi, and
  Jerry~L. Prince, editors, \emph{Simulation and Synthesis in Medical Imaging},
  pages 14--23, Cham, 2017. Springer International Publishing.
\newblock ISBN 978-3-319-68127-6.

\bibitem[Yi et~al.(2019)Yi, Walia, and Babyn]{YI2019101552}
Xin Yi, Ekta Walia, and Paul Babyn.
\newblock Generative adversarial network in medical imaging: A review.
\newblock \emph{Medical Image Analysis}, 58:\penalty0 101552, 2019.
\newblock ISSN 1361-8415.
\newblock \doi{https://doi.org/10.1016/j.media.2019.101552}.
\newblock URL
  \url{http://www.sciencedirect.com/science/article/pii/S1361841518308430}.

\bibitem[Zhu et~al.(2019)Zhu, Bastiaansen, Hakun, Petersson, Wang, and
  Hagoort]{zhu2019100855}
Zude Zhu, Marcel Bastiaansen, Jonathan~G. Hakun, Karl~Magnus Petersson, Suiping
  Wang, and Peter Hagoort.
\newblock Semantic unification modulates n400 and bold signal change in the
  brain: A simultaneous eeg-fmri study.
\newblock \emph{Journal of Neurolinguistics}, 52:\penalty0 100855, 2019.
\newblock ISSN 0911-6044.
\newblock \doi{https://doi.org/10.1016/j.jneuroling.2019.100855}.
\newblock URL
  \url{http://www.sciencedirect.com/science/article/pii/S0911604418301350}.

\bibitem[Zhuang et~al.(2019)Zhuang, Schwing, and Koyejo]{zhuang2019fmri}
Peiye Zhuang, Alexander~G Schwing, and Oluwasanmi Koyejo.
\newblock Fmri data augmentation via synthesis.
\newblock In \emph{2019 IEEE 16th International Symposium on Biomedical Imaging
  (ISBI 2019)}, pages 1783--1787. IEEE, 2019.

\bibitem[Zuo et~al.(2017)Zuo, Fan, Blasch, and Ling]{zuo2017combining}
Haiqiang Zuo, Heng Fan, Erik Blasch, and Haibin Ling.
\newblock Combining convolutional and recurrent neural networks for human skin
  detection.
\newblock \emph{IEEE Signal Processing Letters}, 24\penalty0 (3):\penalty0
  289--293, 2017.

\end{thebibliography}

\end{document}